\documentclass[lettersize,journal]{IEEEtran}
\usepackage{amsmath,amsfonts}
\usepackage{array}
\usepackage[caption=false,font=normalsize,labelfont=sf,textfont=sf]{subfig}
\usepackage{textcomp}
\usepackage{stfloats}
\usepackage{url}
\usepackage{verbatim}
\usepackage{graphicx}
\usepackage{booktabs}
\usepackage{cite}
\usepackage{soul}
\usepackage{multirow}
\usepackage{epstopdf}
\usepackage[hidelinks]{hyperref}
\usepackage{listings}
\usepackage{fancybox}
\usepackage{graphicx}
\usepackage{subcaption}
\usepackage{paralist}
\usepackage{ragged2e}
\usepackage{enumitem}
\usepackage{color}
\usepackage{xcolor,colortbl}
\usepackage{balance}
\usepackage{threeparttable}

\usepackage[framemethod=TikZ]{mdframed}
\usepackage{tcolorbox}
\makeatletter
\newcommand{\mybox}[1]{%
  \setbox0=\hbox{#1}%
  \setlength{\@tempdima}{\dimexpr\wd0+13pt}%
  \begin{tcolorbox}[boxrule=0.5pt, colback=white, arc=4pt,
      left=6pt,right=6pt,top=6pt,bottom=6pt,boxsep=0pt]
    #1
  \end{tcolorbox}
}
\usepackage{tikz}
\usepackage{pgfplots}
\usetikzlibrary{pgfplots.statistics,calc}
\usepackage{color}
\usepackage{xcolor}
\usepackage{array}
\usepackage{amsmath}
\usepackage{centernot}
\usepackage{xspace}
\usepackage{url}
\usepackage{verbatim}
\usepackage{wrapfig}
\usepackage{tabularx}
\usepackage{bbding}
\usepackage{pifont}
\usepackage{wasysym}
\usepackage{makecell}

\makeatletter  
\newif\if@restonecol  
\makeatother  
\usepackage[linesnumbered,ruled,vlined]{algorithm2e}
\usepackage{algpseudocode}  
\newcommand{\tool}{\texttt{PEELING}}
\begin{document}


\title{
Adversarial Testing for Visual Grounding via Image-Aware Property Reduction
}

\author{Zhiyuan Chang, Mingyang Li, Junjie Wang, Cheng Li, Boyu Wu, Fanjiang Xu, Qing Wang
\thanks{Zhiyuan Chang, Mingyang Li, Junjie Wang, Cheng Li, Boyu Wu and Qing Wang are with State Key Laboratory of Intelligent Game, Institute of Software Chinese Academy of Sciences, and University of Chinese Academy of Sciences, Beijing, China. Email: zhiyuan2019@iscas.ac.cn; mingyang2017@iscas.ac.cn; junjie@iscas.ac.cn; licheng221@mails.ucas.ac.cn; boyu\_wu2021@163.com; fanjiang@iscas.ac.cn; wq@iscas.ac.cn.}
}
\markboth{Journal of \LaTeX\ Class Files,~Vol.~14, No.~8, August~2021}
{Shell \MakeLowercase{\textit{et al.}}: A Sample Article Using IEEEtran.cls for IEEE Journals}


\maketitle

\begin{abstract}

Due to the advantages of fusing information from various modalities, multimodal learning is gaining increasing attention. 
Being a fundamental task of multimodal learning, Visual Grounding (VG), aims to locate objects in images through natural language expressions. 
Ensuring the quality of VG models presents significant challenges due to the complex nature of the task.
In the black box scenario, existing adversarial testing techniques often fail to fully exploit the potential of both modalities of information. 
They typically apply perturbations based solely on either the image or text information, disregarding the crucial correlation between the two modalities, which would lead to failures in test oracles or an inability to effectively challenge VG models.  

To this end, we propose {\tool}, a text perturbation approach via image-aware property reduction for adversarial testing of the VG model.
The core idea is to reduce the property-related information in the original expression meanwhile ensuring the reduced expression can still uniquely describe the original object in the image.
To achieve this, {\tool} first conducts the object and properties extraction and recombination to generate candidate property reduction expressions.
It then selects the satisfied expressions that accurately describe the original object while ensuring no other objects in the image fulfill the expression, through querying the image with a visual understanding technique.

We evaluate {\tool} on the state-of-the-art VG model, i.e. OFA-VG, involving three commonly used datasets.
Results show that the adversarial tests generated by {\tool} achieves 21.4\% in MultiModal Impact score (MMI), and outperforms state-of-the-art baselines for images and texts by 8.2\%--15.1\%.
Additionally, by fine-tuning the original model with the adversarial tests, the performance of OFA-VG could be improved by 18.2\%--35.8\% in accuracy.

\end{abstract}



\section{Introduction}
\label{sec:introduction}

Multimodal learning is a machine-learning paradigm that leverages a variety of different data modalities to train the models, including text, images, audio, and so on~\cite{zubatiuk2021development,tsai2019multimodal,gao2020survey}.
With wide applications in daily life, 
multimodal learning has attracted increasing attention in recent years~\cite{zubatiuk2021development,tsai2019multimodal,gao2020survey}.
As a fundamental task in the multimodal learning field, Visual Grounding (VG) aims to locate the object in an image through natural language expressions~\cite{mao2016generation,yang2022improving}, and it affects many downstream tasks such as Visual Question Answering (VQA)~\cite{antol2015vqa, das2018embodied}, 3D Visual Grounding ~\cite{liu2021refer, yuan2021instancerefer}, robot navigation ~\cite{qi2020reverie,anderson2018vision}, autonomous driving ~\cite{kim2019grounding} and photo editing~\cite{cheng2014imagespirit,Turmukhambetov16}.

The reliability of the VG model is of great importance.
For instance, in scenarios where autonomous driving is operated with commands, the VG model is used to understand the meaning of the commands and locate the targets described in the commands. This aids the autonomous driving system in correctly planning its route ~\cite{Thierry20,ChanG022}.
The errors in the VG model could lead to wrong route planning, thereby threatening the safety of autonomous driving, e.g. wrong parking route leads to potential traffic accidents.
Furthermore, the quality issues of the VG are also quite challenging to spot. For instance, Akula et al. \cite{AkulaGAZR20} reported that with certain subtle changes in the linguistic structure of the original expressions, the performance of the VG models will obviously decrease.
As a vital means of quality assurance, the automatic testing techniques for VG  models have not been well studied, and are badly desired in real-world practice.

In industrial scenarios, where testers do not have access to the internal parameters of the model.
 In this case, adopting white-box approaches is much more difficult, while the black-box approach becomes a more practical option.
Under the black box scenario, many adversarial approaches test the image-oriented or text-oriented deep learning models by bringing a tiny perturbation for the inputs which keeps the original output oracle unchanged. 
Given the two types of input (i.e., images and expressions) in the VG models, existing techniques can only carry out the perturbation either solely based on the image or solely on the natural language expression. 
These single modality-oriented adversarial techniques are not suited for VG models, which are designed for multimodal scenarios, due to the following two aspects.

\textbf{First}, for the images, traditional perturbation techniques usually introduce a certain degree of random noise to the whole images at different granularity levels, i.e. pixels and blocks, or conduct unexplained image synthesis~\cite{HendrycksD19,GeirhosRMBWB19,ZhuPIE17}.
In particular, they lack the ability to accurately determine the magnitude of the perturbation's effect on the target objects in the image.
Based on the above implementation, some image features that significantly impact the object localization will undergo unpredictable changes before and after the perturbation (details are discussed in Section \ref{sec:analysis_img}), which leads to the failure of test oracles (i.e., the expected output inconsistent with the test oracle).

\textbf{Second}, for natural language texts, most of the existing testing approaches (e.g., Q\&A system testing and translation system testing) conduct semantically-equivalent perturbations~\cite{SunZHPZ20,0004Z0HP022,RusertSS22}.
Specifically, given the original texts, common perturbations include synonym substitution, back translation, simulating some character-level spelling errors, and removing some uninformative words (e.g., adverbs and conjunctions) to maintain semantic consistency before and after perturbations~\cite{0002LSBLS18,LiuF021}.
Due to a lack of perception of core semantic components present in both expressions and images, these techniques have limitations in such multimodal scenarios and the perturbations are relatively subtle and shallow.
Accordingly, the adversarial tests hardly challenge the VG models and therefore cannot effectively detect issues.
For the VG task, since both image and natural language expression are intertwined and interact with each other, we aim to further exert the perturbation potency by considering these two modalities of information.

\begin{figure}[h]
\centering
\includegraphics[width=8cm,height=2.8cm]{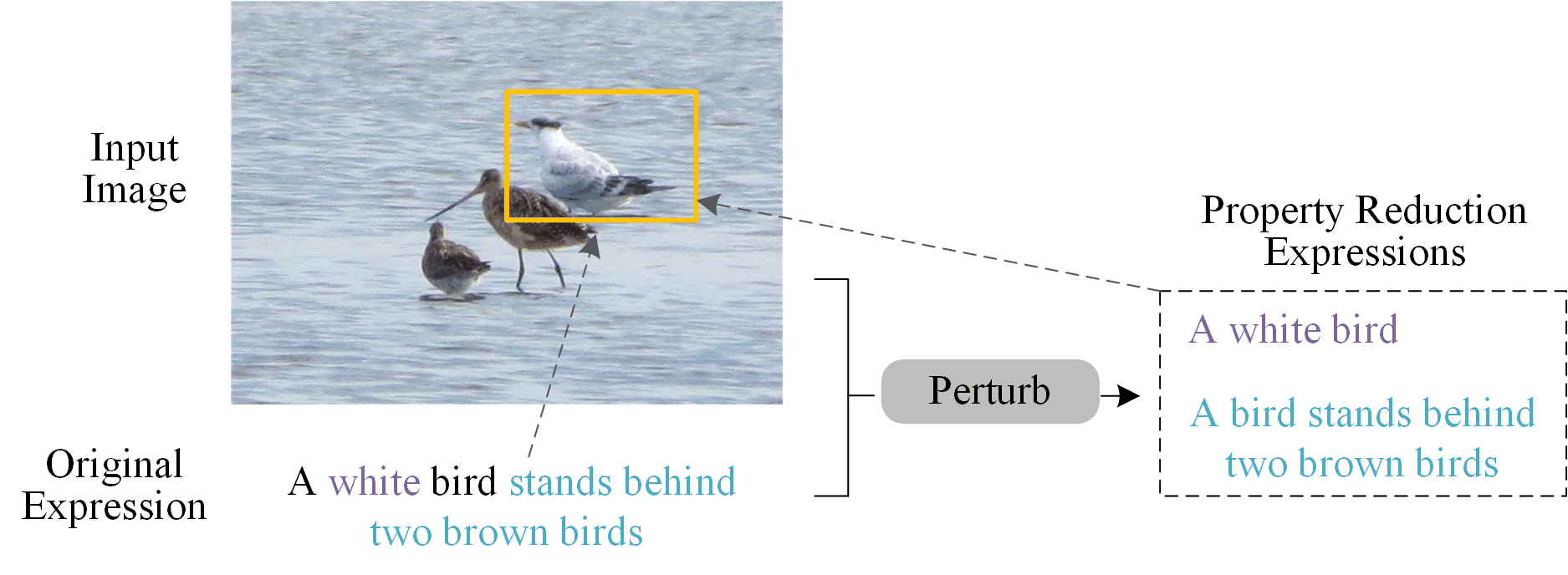}
\caption{An example of perturbation via property reduction for VG}

\label{fig:motivation}
\end{figure}


In the VG task, it is common for expressions to include an abundance of properties when describing an object. In such cases, even if we remove certain redundant properties from the original expressions (i.e., through property reduction), the target object can still be identified within the image and remains localized in the same region as before the property reduction was applied.
Figure \ref{fig:motivation} demonstrates an example for detailed illustration.
With the original expression ``A white bird stands behind two brown birds'', the VG model aims at locating the object framed by the rectangle in the image.
By analyzing the semantic elements, the original expression describes a ``bird'' and two associated properties, i.e., the color property ``white'' and the position property ``stands behind two brown birds''.
By the image understanding, it becomes apparent that either of these properties is redundant, and even after removing either property, the target object can still be accurately located.
By performing property reduction, the original expression can be transformed into two valid expressions: ``A white bird'' or ``A bird stands behind two brown birds.'' These modified expressions effectively convey the necessary information while eliminating redundant properties.
Based on Shannon's Entropy theory~\cite{shannon2001mathematical}, the smaller the amount of information an expression contains, the higher the uncertainty is with the expression.
Accordingly, the VG models might show low confidence for the expressions after property reduction. 
Taken in this sense, this paper aims to find the suitable property to reduce and derive the property reduction expression, so as to explore whether the VG model exhibits correct behavior when faced with such modified expressions. 

The biggest challenge to performing property reduction lies in ensuring that the expression accurately describes the target object and that no other objects in the image fulfill the given expression, i.e., the test oracle (target region) remains unchanged.
By referring to the object together with its properties, we can generate multiple candidate property reduction expressions.
We can then leverage the semantics of the image to determine whether a candidate expression uniquely characterizes the target object, which can make better use of the two modalities of information in the multimodal model.

In this paper, we propose a text perturbation approach via image-aware property reduction (named {\tool}) for adversarial testing of VG.
{\tool} generates new expressions with reduced properties while keeping the target object in the image unchanged.
Specifically, through object and properties extraction, candidate expression generation by object and property recombination, and image-aware expression selection, {\tool} perturbs the original expressions with redundant properties resulting in property reduction expressions that can still locate the original target region in the image.
Additionally, {\tool} further conducts semantically-equivalent perturbations at the character/word/sentence level.
This can diversify the range of perturbation operations, so as to facilitate uncovering more issues present within the VG model. 
Subsequently, it generates the final adversarial expressions for VG testing.

We evaluate {\tool} on OFA-VG, a state-of-the-art VG model, 
together with three commonly used VG datasets, i.e. RefCOCO~\cite{YuPYBB16}, RefCOCO+~\cite{YuPYBB16} and RefCOCOg~\cite{MaoHTCY016}.
Results show that, in issue detection ability, {\tool} achieves 21.4\% in MultiModal Impact score (MMI), and outperforms state-of-the-art baselines for images and texts by 8.2\%--15.1\%.
In addition, the two perturbations in {\tool}, i.e. Property Reduction Expression Perturbation and Semantically-Equivalent Expression Perturbation, could respectively contribute to 11.1\% and 14.2\% MMI on average. 
Furthermore, the performance of the VG model can be enhanced (accuracy increases by 18.2\%--35.8\%) by fine-tuning it using the adversarial tests generated by {\tool}.
The key contributions of this paper are as follows:
\begin{itemize}
    \item
    We propose a text perturbation approach for adversarial testing of VG based on image-aware property reduction.
    This is the first approach for adversarial testing the VG model considering the two involved modalities (image and text), to the best of our knowledge, and can motivate other testing practice related to multimodal learning models. 
    
    \item 
    We perform comprehensive experiments to evaluate the effectiveness of {\tool}, the experimental results show that {\tool} exhibits remarkable issue detection ability and significantly outperforms the state-of-the-art baselines on three widely-used datasets.
    
    \item
    By fine-tuning the original model with the adversarial tests, the performance of the VG model could be significantly improved.
   
    \item
    We provide the public reproduction package\footnote{https://github.com/IMGAT/PEELING} including the tool implementation and datasets.
\end{itemize}

\section{Background and Motivation}
\label{sec:background}

\subsection{Visual Grounding}
\label{sec:backgroundone}

VG is to localize the objects in the image according to a specific expression in natural language. 
Taking Figure \ref{fig:motivation} as an example, given an input image and an expression ``A white bird stands behind two brown birds'', the VG model will output an image with the located target framed by a rectangle.
With the development of VG models,  pre-trained models have become the mainstream approach to this task. 
Under this paradigm, there are several representative models such as ViLBERT~\cite{lu2019vilbert}, LXMERT~\cite{TanB19}, and OFA-VG~\cite{wang2022ofa}.
Among these models, the OFA-VG model has made significant strides, achieving state-of-the-art results in VG, and is widely downloaded and employed from publicly available pre-trained model hubs\footnote{https://huggingface.co/}.
For the datasets, RefCOCO, RefCOCO+, and RefCOCOg are widely-used by researchers under the VG task due to the diverse range of the scenarios and the high quality of the annotations.
In our study, we select the state-of-the-art model, OFA-VG, as the testing subject and the datasets RefCOCO, RefCOCO+, and RefCOCOg for test generation and evaluation.

\subsection{Motivation}
\label{sec:motivation}

\begin{figure}[h]
\centering
\includegraphics[width=\linewidth]{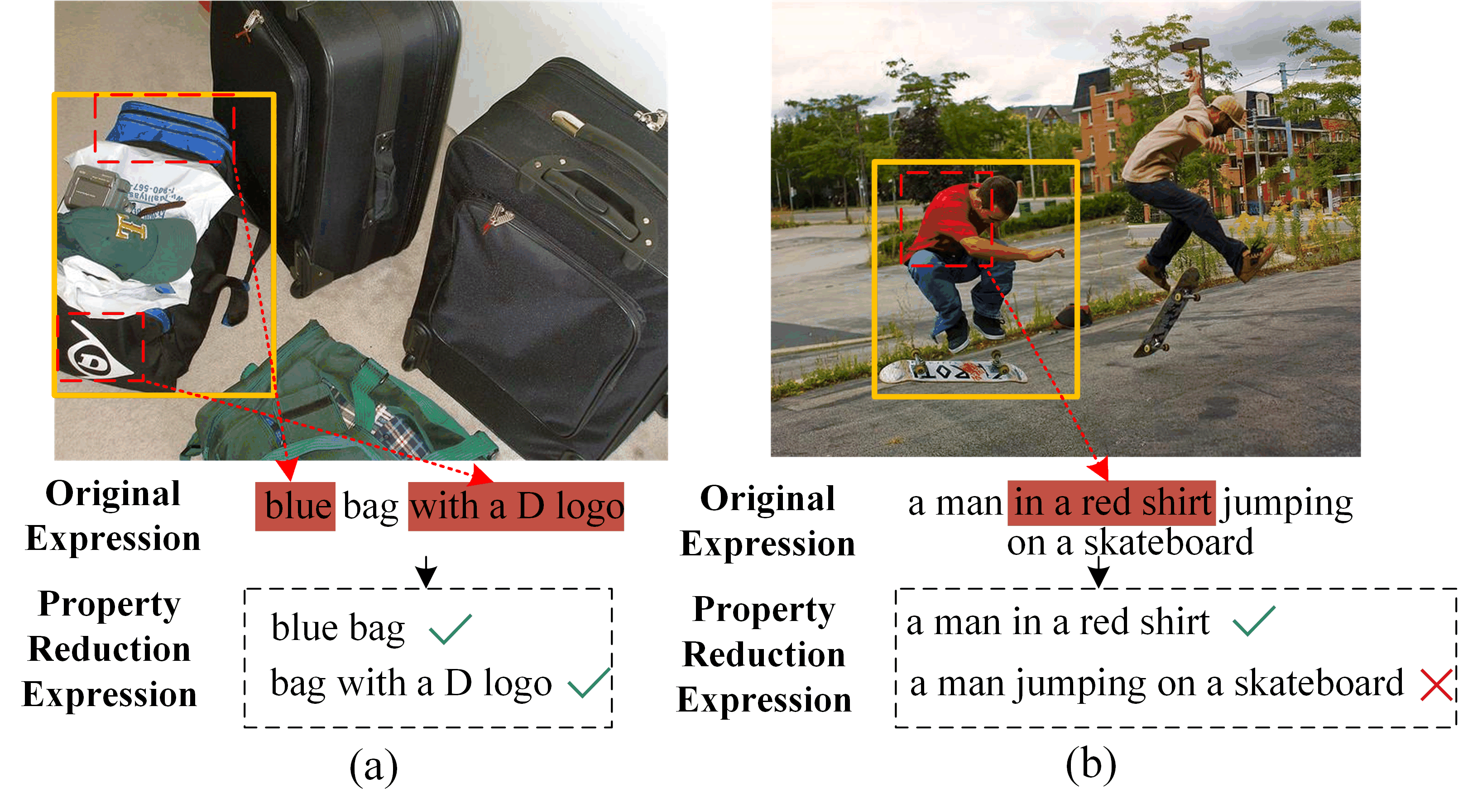}
\caption{Two examples to demonstrate the feasibility and the challenge of property reduction}
\label{fig:background_motivation}
\end{figure}

On the one hand, we explore the feasibility of property reduction for the original expression. 
Generally, a target object in an image typically contains multiple properties, and the expressions contain multiple linguistic components to describe corresponding properties in many cases.
In such cases, the expressions typically contain abundant properties, and it provides us with the possibility to generate new tests from the original expression through property reduction.
Take the subfigure (a) in Figure \ref{fig:background_motivation} as an example.
Through the image understanding, the target object ``bag'' distinguishes from the others by either of the two explicit properties, i.e., the color ``blue'' and the decorative pattern ``with a D logo''.
Therefore, given an original expression ``blue bag with a D logo'', removing either property could form a candidate property reduction expression for VG testing.

On the other hand, although there are multiple properties of the target object in the expression, it does not mean that they can be arbitrarily recombined when conducting expression perturbation.
Take the subfigure (b) in Figure \ref{fig:background_motivation} as an example.
Given the expression ``a man in a red shirt jumping on a skateboard'', it describes two properties of the target object ``a man'', i.e., ``in a red shirt'' and ``jumping on a skateboard''.
By the image understanding, we find that the two properties play completely different roles in the VG task.
Specifically, the single property ``jumping on a skateboard'' could not accurately locate the target since there are two men with the same property, while the property ``in a red shirt'' is discriminative.
Therefore, of the two expressions, only the one ``a man in a red shirt'' could accurately locate the original object in the image.
In general, it is an appropriate path to perturb the original expression through property reduction, but it is a prerequisite and challenging to determine the impact of different properties in the VG task.

\section{Approach}
\label{sec:approach}

\begin{figure*}[h]
  \center{\includegraphics[width=\linewidth]{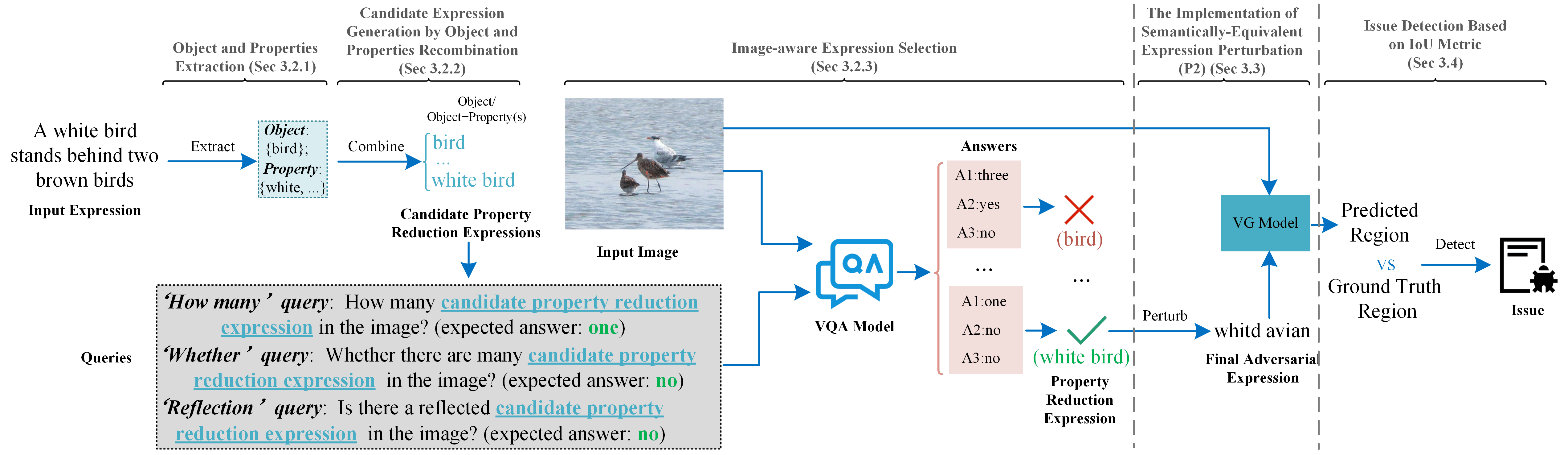}}
    \caption{
     The overview o f {\tool}
     }
    \label{fig:MR1_process}    
\end{figure*}

The core idea of {\tool} is that the VG model should locate the original target object in the image if {\tool} can find a property reduction expression that describes the original object and there are no other objects that satisfy the description. 
To achieve this, {\tool} first extracts the object and its properties in the original expression and generates several candidate property reduction expressions through the object and its associated properties recombination.
Then, {\tool} validates whether a candidate property reduction expression is suitable by understanding the semantics of the image, i.e., querying the image with the VQA model.
Furthermore, {\tool} performs semantically-equivalent perturbations to further enhance its ability in issue detection.
Figure \ref{fig:MR1_process} shows the overview of {\tool}.
In the following subsections, we will first describe two perturbations (Section \ref{approach_MR}).
Next, we present details of implementing perturbation 1 (P1) (Section \ref{approach_candidate}) and perturbation 2 (P2) (Section \ref{sec:Semantically-equivalent}), and the process of detecting issues for the VG model (Section \ref{approach:issuedetect}).

\subsection{Perturbations}
\label{approach_MR}
{\tool} proposes two perturbations to generate the tests for the VG model by combining image and text information.
The two perturbations both perturb the input expression, donated as $e$, and the perturbed expression is denoted as $e'$.
For simpler presentation, the regions in the images corresponding to the expressions before and after the perturbation are $R(e)$ and $R(e')$, respectively.

\textbf{P1: Property Reduction Expression Perturbation.}
The assumption of designing this perturbation is that if we could find a property reduction expression $e'$ that describes the original object in the image and there are no other objects in the image that satisfy the description.
The region $R(e')$ should be consistent with $R(e)$.

\textbf{P2: Semantically-Equivalent Expression Perturbation.}
The assumption of designing this perturbation is that we could find a semantically-equivalent expression $e'$ to the original expression, thus further enhancing the diversity of the adversarial tests.
Since the expression semantics remain unchanged, the region $R(e')$ should be consistent with $R(e)$.

\subsection{The Implementation of Property Reduction Expression Perturbation (P1)}
\label{approach_candidate}

In order to implement property reduction expression perturbation, {\tool} undergoes three steps:
(1) Object and Properties Extraction, where {\tool} extracts object and properties from the original expression; (2) Candidate Expression Generation by Object and Properties Recombination, where {\tool} constructs candidate property reduction expression by combining object and its associated properties; (3) Image-aware Expression Selection, where {\tool} validates whether candidate property reduction expressions keep the test oracle unchanged and selects the satisfactory ones as tests.
The following introduces the details of three steps.


\subsubsection{\textbf{Object and Properties Extraction}}
\label{approach_extraction}


By analyzing the semantic elements within the expressions, we find the containing ``object'' and its associated ``properties'' (the definitions and examples are given in Table \ref{tab:entity_categories}) take the prominent effects on the VG task from the cognitive view.
In order to conduct property reduction, it is a prerequisite to first extract the object and properties within the expression.
Inspired by the emergence ability and widespread application of Large Language Model (LLM), {\tool} employs ChatGPT\footnote{https://openai.com/blog/chatgpt}, which is a dominated LLM in practice, for unsupervised object and properties extraction.
Recently, it has been demonstrated that ChatGPT, in particular, possesses considerable text understanding capabilities~\cite{Zhong2023ChatGPT} and achieves impressive performance on the information extraction task~\cite{Bo2023entity}.

\begin{table}[h]
\caption{The definitions and examples of object and property for the expressions}
\label{tab:entity_categories}
\resizebox{\columnwidth}{!}{
\begin{tabular}{ccc}
\toprule
\textbf{Category}    &   \textbf{Definition} &   \textbf{Example}  \\   
\midrule
Object  &  the target to be located in the image  &  ``bird'' in the original expression of Figure \ref{fig:motivation} \\
Property   &  the modified expression for the object  &   ``white'' and ``stand behind two brown of birds'' \\
& & in the original expression of Figure \ref{fig:motivation}     \\

\bottomrule
\end{tabular}
}
\end{table}

\begin{figure}[h]
\centering
\includegraphics[width=0.5\textwidth]{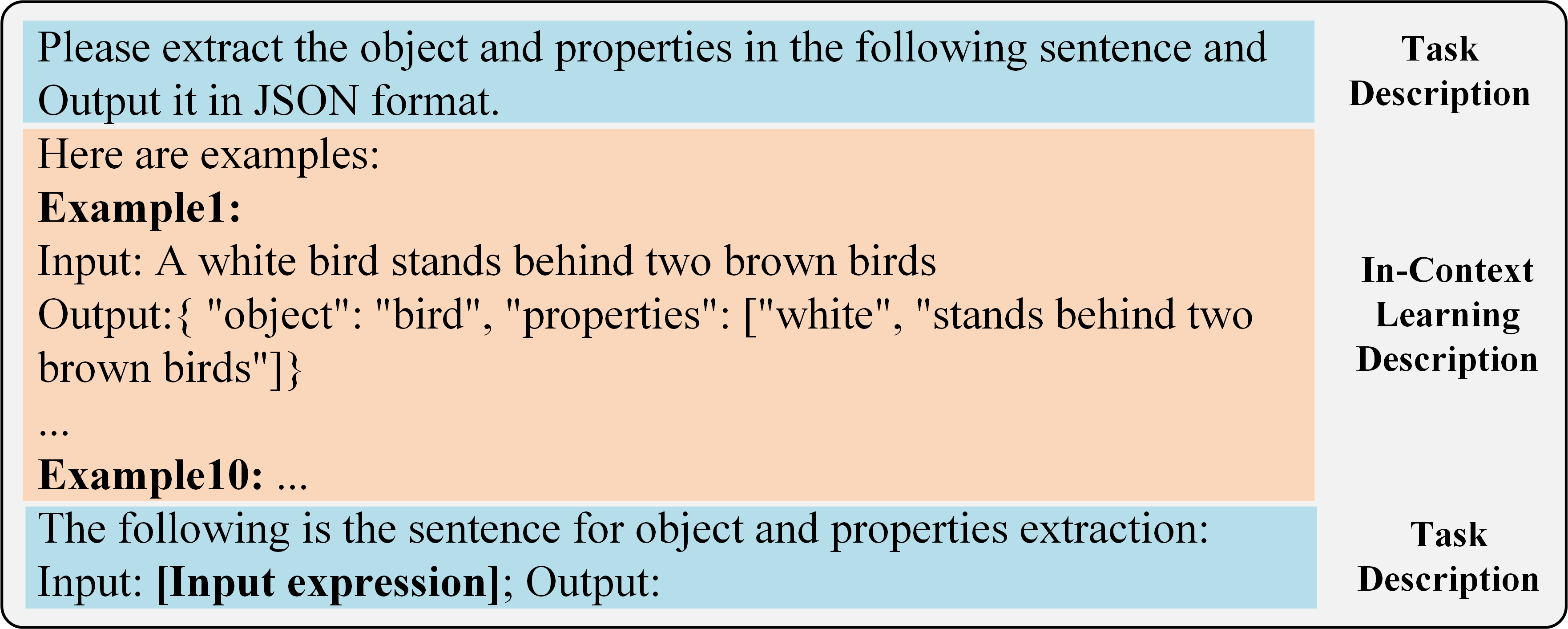}
\caption{Example of the prompt template for object and properties extraction}
\label{fig:prompt_example}
\end{figure}

To employ ChatGPT for the information extraction task, it's crucial to craft suitable questions, i.e., prompts.
The prompt template designed by the {\tool}, as shown in Figure \ref{fig:prompt_example}, consists of two parts of the elements, i.e., task descriptions and in-context learning description.
Task descriptions aim to help ChatGPT understand instructions and standardize the form of outputs, and they are created by rewriting the prompt template in the previous study in information extraction with ChatGPT by Li et al.~\cite{Bo2023entity}.
In-context learning description is, in essence, a collection of input-output samples that guide the model's learning process.
The assumption behind introducing the description is that incorporating the samples containing task-specific knowledge could enhance the performance of ChatGPT when solving the problem with its own peculiar characteristics~\cite{BrownMRSKDNSSAA20}.

To obtain the samples for in-context learning, {\tool} first selects expressions whose length is beyond the average text length of the expressions in the candidate set, i.e. the training set in the three datasets (details in Section \ref{sec:exeperiment_dataset}).
The intuition behind the criteria is that the complexity of an expression increases with its length.
Simultaneously, {\tool} intentionally exposes ChatGPT to more complex samples, under the assumption that a model grasping complex cases will naturally perform well with simpler ones.
After that, we manually labeled selected expressions and formulated them into the in-context learning description within the prompt in question-and-answer forms.

During the application stage, the ``[Input expression]'' placeholder within the prompt is replaced with the actual input expression. Subsequently, ChatGPT generates extraction results that are formatted to correspond closely with our predetermined sample output.

\subsubsection{\textbf{Candidate Expression Generation by Object and Properties Recombination}}
For an original expression, this step aims to generate candidate property reduction expressions that carry less but sufficient properties.
Specifically, given the extracted object and properties, {\tool} combines them according to the following two strategies.

\textit{\textbf{Strategy-1}}: The object is used independently (i.e., without any properties) as a candidate property reduction expression.
For an image, there might exist distinguishing differences between the target object and the other objects in terms of category, and the target object could be located without adding extra properties. 
For example, in an image with only one man and a group of women, the object ``man'' in the expression ``The man surrounded by women'' is sufficient to locate the region of the man in the image.

\textit{\textbf{Strategy-2}}: Connect the object and the properties according to their position in the original sentence. 
To generate a fluent expression, {\tool} follows the original positions of object and properties in the original expression when connecting them.
For example, when connecting the object ``bird'' and the property ``stands behind two brown birds'' in the expression ``a bird stands behind two brown birds'', they should be connected in accordance with their positions in the original expression.
Otherwise, the resulting sentence is ungrammatical and not fluent, such as ``stands behind two brown birds bird''.
 
Please kindly note that for an expression, {\tool} would not concatenate the object and all its associated properties, since it violates the intention of reducing the properties from the original expression. 
Based on these two strategies, {\tool} obtains candidate property reduction expressions that are likely to locate the original target region in the image.

\subsubsection{\textbf{Image-aware Expression Selection}}
This step aims to validate whether the properties conveyed by each candidate expression are sufficient to locate the original target object and select satisfied ones for VG testing.
In this process, {\tool} takes into account the content of the image.
By leveraging Visual Question Answering (VQA) technology, {\tool} designs three queries for a VQA model to obtain the model responses.
Based on the responses to these queries, {\tool} validates whether each candidate is sufficient to locate the original target object in the image.


\textbf{\textit{`How Many' query. }}
This query aims at determining how many described objects exist in the image, and the expected answer of this query is one. 
If the answer is larger than one, there would be a very high possibility that the candidate property reduction expression would not uniquely describe the original target object/region. 

\textbf{\textit{`Whether' query. }}
This query aims at knowing whether there is more than one described object existing in the image, and the expected answer to this query is no.
It serves a similar purpose as the `How many' query, but from a different perspective.  
They both aim to determine whether there are multiple objects in the image that match the candidate property reduction expression.
If the answer is yes, there is a high probability that the candidate property reduction expression would not uniquely describe the target object/region.

\textbf{\textit{`Reflection' query. }}
This query aims to know whether many described objects being reflected appear in the image (e.g. mirror reflections).
Since the reflected objects are not real, the expected answer to this query is no.
If the answer is yes, there is a high probability that a candidate property reduction expression refers to the target in the reflection and would not uniquely describe the target object/region.

For a candidate property reduction expression, {\tool} employs OFA-VQA~\cite{wang2022ofa}, which is the state-of-the-art VQA model, to obtain the corresponding answers to the three queries, and adopts an ``AND'' logic-based strategy to select the satisfied property reduction expressions.
Specifically, only when the answers to all three queries meet expectations do we consider it a satisfied expression to be used for subsequent issue detection.
This strategy could mitigate the risk that the final selected expression would not be subject to errors resulting from a single query, and could improve the quality of the adversarial tests as well as the reliability of {\tool}.

\subsection{The Implementation of Semantically-Equivalent Expression Perturbation (P2)}
\label{sec:Semantically-equivalent}
To further improve the diversity of the adversarial tests, {\tool} introduces the semantically-equivalent text perturbation techniques to the property reduction expressions.
Traditional techniques conduct perturbations on the original text under the premise of keeping the test oracle unchanged, and the perturbations could commonly be classified into three levels, i.e., character-level, word-level, and sentence-level~\cite{LiuFYS0X22}.
According to the three levels, {\tool} respectively employs one of the representative techniques to perturb the property reduction expressions.
For character-level perturbation, {\tool} selects the keyboard mistake~\cite{LiuF021}, which is the spelling mistake caused by wrongly pressing similar keys on the keyboard.
In detail, {\tool} operates by selecting one word in the expression at random, and then perturbs a single character within that chosen word.
For word-level perturbation, {\tool} selects the synonym substitution~\cite{LiuFYS0X22}, which replaces the original word with a synonym.
In detail, {\tool} operates by selecting one word in the expression at random, and then perturbs the chosen word.
For sentence-level perturbation, {\tool} employs the back translation technique~\cite{LiuFYS0X22}, which translates the original sentence into another language and then back into English.
In detail, {\tool} translates the expression first into German and then from German back into English.
The above three perturbations are all implemented by the nlpaug toolkit\footnote{https://github.com/makcedward/nlpaug}.
For a property reduction expression, {\tool} applies a combination of three semantically-equivalent perturbations to the property reduction expressions, first at the sentence level, followed by the word level, and finally at the character level.

After the three-level perturbations, {\tool} obtains the final adversarial expressions as the test inputs together with their paired original image. 
Each test input and its test oracle are fed into the VG model for issue detection.

\subsection{Issue Detection Based on IoU Metric}
\label{approach:issuedetect}
Given a final adversarial expression and its paired image, this phase produces the predicted region within the image and detects the issues by comparing the predicted region and test oracle.
To accomplish this, {\tool} initially calculates the overlap between the predicted region and the test oracle using the Intersection over Union (IoU) metric~\cite{EveringhamGWWZ10}. 
IoU is a measure that quantifies the similarity between two regions by comparing the area of their intersection with the area of their union and is commonly used to measure the performance of the VG models.
The IoU metric is calculated using the following formula:
\begin{equation}
IoU(A, B) = \frac{Area(A \cap B)}{Area(A \cup B)}
\end{equation}
where $A$ represents the predicted region, $B$ represents the ground truth region, $A \cap B$ is the intersection of the two regions, and $A \cup B$ is their union. 
The IoU value ranges from 0 to 1, where 0 indicates no overlap and 1 represents a perfect match between the predicted and ground truth regions.
In general, the IoU threshold 0.5 serves as a standard criterion to evaluate the performance of VG models~\cite{ChenZSWL23,YangKDO23}.
In our study, {\tool} follows the IoU threshold of 0.5.
Given a predicted region, if the IoU value is greater than 0.5 compared with the ground truth region, it indicates that the predicted region has a substantial overlap with the ground truth region and is considered a correct prediction.
On the contrary, if the IoU value is less than or equal to 0.5 compared with the test oracle, the corresponding sample is regarded as an issue sample.

\section{Experiment}
\label{sec:experiment}
\subsection{Research Questions}

\textbf{RQ1: Can the adversarial tests generated by {\tool} effectively detect issues for the VG model?}
We explore the performance of {\tool} in two aspects, i.e., the correct rate of the adversarial tests and the issue detection ability.

\textbf{RQ2: Can the two perturbations improve the issue detection ability of {\tool} for the VG model?}
We conduct ablation experiments to evaluate the effectiveness of the two perturbations in issue detection.

\textbf{RQ3: Can the adversarial tests generated by {\tool} help improve the performance of the VG model?}
We investigate whether the performance of the VG model could be improved with the adversarial tests generated by {\tool}.

\textbf{RQ4: Can {\tool} effectively extract objects and properties from the VG expressions?
}
We explore the effectiveness of {\tool} in extracting objects and properties from the expressions, which serves as the basis for P1.

\subsection{Dataset}
\label{sec:exeperiment_dataset}
To evaluate the performance of {\tool}, we employ three commonly used public VG datasets, i.e. RefCOCO, RefCOCO+, RefCOCOg.
Specifically, RefCOCO contains a training set, a validation set, and a test set with 120,624, 10,752, and 10,834 samples (i.e., image and expression pair) respectively, where the training set and validation set are used to optimize the model hyper-parameters during the training phase, and the test set is used to assess its performance after the model is finalized in the VG task.
RefCOCO+ contains a training set with 120,191 samples, a validation set with 10,759 samples, and a test set with 10,615 samples.
For RefCOCOg, it is not publicly partitioned.
We follow the partition approach in the previous study~\cite{NagarajaMD16} and obtain a training set with 80,521 samples, a validation set with 4,896 samples, and a test set with 9,602 samples.
Following the previous studies \cite{AkulaGAZR20,Shen0ZW0T22}, we leverage the test sets in the three datasets as the input of {\tool} to generate new tests.

As for the required input, {\tool} relies on the expression that has redundant properties to generate new tests.
Therefore, the proportion of such expressions in the original dataset is an important factor affecting the applicability of {\tool}.
To this end, for each of the three datasets, we randomly sampled 500 samples from each of the three test sets and manually inspected each property with the expressions.
If any redundant property exists in a sample (i.e., the expression could still manually locate the object in the image after removing the property), it is considered a required expression.
Finally, we found that 46.2\% sampled expressions have redundant properties on average, which indicates that required expressions are prevalent in three datasets.



\subsection{Testing Subject}
To investigate the performance of {\tool} in VG testing, we introduce the state-of-the-art VG model, OFA-VG, as the testing subject.
As reported in the previous study~\cite{wang2022ofa}, OFA is a unified sequence-to-sequence pre-trained model that unifies modalities (i.e. vision and language) and tasks.
 Therefore, after fine-tuning OFA on various multimodal datasets, it can be employed in a wide range of multimodal scenarios,  such as VG, VQA, and Image Captioning (IC).
OFA-VG, finetuning OFA on the VG dataset, has achieved promising performance in RefCOCO, RefCOCO+ and RefCOCOg.
In our study, we use the public OFA-VG model\footnote{https://github.com/OFA-Sys/OFA} trained by the previous study as the testing subject.


\subsection{Experiment Design}
\label{sec:experiment_design}

To answer RQ1, for each dataset, {\tool} generates the adversarial tests using the samples in the test set, and the adversarial tests are sent to the OFA-VG model for issue detection.
First, we evaluate the correct rate of the adversarial tests by manual checking.
Specifically, for each dataset, we randomly sample 100 adversarial tests and manually inspect the correct rate of the adversarial tests.
To ensure the quality, we built a labeling team including a senior researcher and two Ph.D. students for the inspection.
Once different opinions arise, the labeling results are determined by the team discussion.
For a test, if the three members generally agree that the final adversarial expressions could still locate corresponding objects in images, it is considered a correct one.
Otherwise, it is considered a wrongly one.
After that, we calculate the ratio of correctly adversarial tests to all adversarial ones.
The above steps are repeated three times and the average value of the three ratios is taken as the final correct rate of the adversarial tests.
Second, we evaluate the issue detection performance of {\tool}.
For comparison, we set up the two groups of experiments, i.e.,  (1) \textbf{Original}: The performance of OFA-VG on the original test set;
and (2) \textbf{{\tool}}: The performance of OFA-VG on the adversarial tests generated by {\tool}.
After that, we investigate the relative performance decrease of OFA-VG on the adversarial tests and original test set, indicating the issue detection ability of {\tool}.
Moreover, we introduce two state-of-the-art text perturbation approaches and two state-of-the-art image perturbation approaches (illustrated in Section \ref{sec:exe_baselines}), as the baselines.

To answer RQ2, we conduct ablation experiments to explore the effectiveness of the two perturbations in {\tool}.
Specifically, we conduct experiments under two experimental settings by removing the perturbation(s) in terms of ${\tool} \setminus X$, where X is either P1 or P2.
Using the relative performance decrease of {\tool} as the baseline, we explore whether there is a significant reduction in the relative performance decrease after removing P1 and P2 respectively.

To answer RQ3, we adopt the fine-tuning strategy to explore whether the performance of OFA-VG could be enhanced with the adversarial tests generated by {\tool}.
Following the experimental setting in the previous study~\cite{ChenJX21}, we divide the adversarial tests into a training set and an evaluation set in the ratio of 8:2 by random sampling.
After that, we fine-tune the OFA-VG model using the training set and evaluate the performance on the evaluation set.
This setting is commonly employed to enhance the performance of deep learning models in previous studies~\cite{Gupta20, HeMS20}.
For comparison, we also obtain the performance of the original OFA-VG model before fine-tuning.
To avoid randomness, each experiment is repeated five times, and the average is considered the final result.

To answer RQ4, we investigate the performance of {\tool} in extracting objects and corresponding properties.
To achieve the in-context learning, we first select 10 expressions from the training set of the three datasets according to the sampling strategy in Section \ref{approach_extraction}.
For each expression, we manually label the object and properties in the expressions by the labeling team. 
For an expression, if all three members agree on object and properties labeling results, it is used as a ground truth.
Once different opinions arise, the labeling results are determined by the team discussion.
Finally, we obtain 10 expressions (involving 10 objects and the corresponding 13 properties) as the in-context learning samples\footnote{The details of the 10 labeled expressions are provided in our public package.}.
To investigate the advantages of {\tool}, we introduce the state-of-the-art information extraction model (details in Section \ref{sec:exe_baselines}) for comparison.
The experiment is repeated five times, and the average value is considered the final performance.


\subsection{Baselines}
\label{sec:exe_baselines}
In order to investigate the advantages of {\tool}, under the black box framework, we introduce two state-of-the-art text adversarial testing approaches and two state-of-the-art image adversarial testing approaches.

\textbf{CAT~\cite{0004Z0HP022}}:
It is the state-of-the-art text adversarial testing approach for the translation system under the framework of black box.
Specifically, it uses a BERT model~\cite{DevlinCLT19} to encode the sentence contexts and generate the candidate words for the position being replaced.
After that, CAT selects the substituted sentences that are similar to the original ones and uses them for the Translation system testing.
In our study, we leverage the word substitution technique\footnote{The implementation is publicly provided at https://github.com/zysszy/CAT.} to perturb the expressions in the VG task.


\textbf{QAQA~\cite{Shen0ZW0T22}}:
It is the state-of-the-art text adversarial testing approach for the question-answering system under the framework of black box.
The core idea in QAQA is to add redundant sentences to the query and context.
Given an original query, QAQA identifies similar queries, adds them to the original query, and utilizes the concatenated queries to test the question-answering system.
In our study, reuse its public package\footnote{https://github.com/ShenQingchao/QAQA} for reproduction.
For each expression to be transformed, we retrieve similar queries within the training set in the datasets and construct the concatenated queries for VG testing.


\textbf{SIN~\cite{GeirhosRMBWB19}}:
It is the state-of-the-art image adversarial testing approach for object recognition models under the framework of black box.
Specifically, SIN uses neural networks to separate and recombine content and style from different images, creating a new image that combines the content of one image with the artistic style of another.
In our study, we use SIN to perturb the input images and reuse the package provided by the paper \footnote{https://github.com/rgeirhos/Stylized-ImageNet} for reproduction.

\textbf{Feature-based Image Perturbation (FIP)~\cite{HendrycksD19}}:
It is the state-of-the-art image adversarial testing approach for object recognition models under the framework of black box.
FIP perturbs the pixel values of an image by incorporating noise, blur, weather, and digital effects. 
It involves a total of 15 distinct perturbation modes.
In our study, we use FIP to perturb the input images and reuse the package provided by the paper \footnote{https://github.com/bethgelab/imagecorruptions} for reproduction.

In order to investigate the advantages of the extraction performance of {\tool}, we introduce a state-of-the-art supervised model as a baseline.

\textbf{Tscope~\cite{ChangLWWL22}}:
It is the state-of-the-art information extraction approach used in SE communities.
It first retrieves the candidate entities utilizing the span strategy. 
These candidate entities are then embedded into hidden representation through the BERT language model. 
Finally, the corresponding hidden representation of each entity is input to the linear classifier to be categorized into different categories.
In our study, we use Tscope to extract objects and properties and reuse the package provided by the paper\footnote{https://github.com/czycurefun/testcase\_detection} for reproduction.

Since the supervised model requires labeled samples for training.
We select 400 samples using Stratified Sampling from the training sets in the three VG datasets for labeling.
For each expression, we label the objects and properties following the same steps as labeling in-context learning input-output samples.
Finally, we obtain 400 expressions (involving 400 objects and the corresponding 575 properties) as the ground truth to evaluate the performance of the object and properties extraction model.
With the 400 labeled expressions, we divide them into a training set and an evaluation set in the ratio of 8:2 by random sampling.
For each experiment, we train the supervised model using the labeled expressions in the training set and evaluate the extraction performance on the evaluation set.
The object and properties extraction performance of the {\tool} is also evaluated on the evaluation set.


\subsection{Evaluation Metrics}
We introduce the evaluation metrics from four aspects: (1) the performance of the VG task, (2) the quality of the adversarial tests, (3) the issue detection ability of the adversarial tests, and (4) the performance of object and properties extraction.
To measure the performance of the VG task, we use \textit{Accuracy (ACC)}, the ratio of correct predictions to all predictions, which is the commonly-used metric in the VG task~\cite{MaoHTCY016,YangXYLLH22}.
To measure the quality of the adversarial tests, we use the \textit{Adversarial Tests Correct Rate (ATCR)}, the ratios of the correctly adversarial tests to all the adversarial tests.
To measure the issue detection ability of the adversarial tests, we introduce \textit{MultiModal Impact score (MMI)}, which is a commonly used metric to measure the issue detection ability in terms of the performance differences across different test set~\cite{Qiu2022}.
MMI refers to the averaged accuracy decrease of the VG model on adversarial tests compared with the accuracy on the original test set, i.e., $MMI = ({A_{o} - A_{a}})/{{A_{o}}}$, where $A_{p}$ is the accuracy on the adversarial samples and $A_{o}$ is the accuracy on the original samples.


For the object and properties extraction performance, we introduce three widely-used metrics in information extraction, i.e., Precision, Recall, and F1~\cite{EbertsU20,LampleBSKD16,LuanWHSOH19}.
(1) \textit{Precision}, which refers to the ratio of the number of correct predictions to the total number of predictions; 
(2) \textit{Recall}, which refers to the ratio of the number of correct predictions to the total number of samples in the ground truth; 
(3) \textit{F1-Score}, which is the harmonic mean of precision and recall.

\section{Results}
\label{sec:result}
\begin{table*}[t]\huge
  \caption{The correct rate and the issue detection ability of adversarial tests generated by {\tool} and baselines}
  \label{tab:link_performance}
  
\resizebox{\textwidth }{!}{
\begin{threeparttable}
\begin{tabular}{cc|cccccccccccccccccccc}
\toprule
\multirow{2}{*}{\textbf{Dataset}} &
  \multirow{2}{*}{\textbf{Metric}} & 
  \multirow{2}{*}{\textbf{Original}} &
  \multirow{2}{*}{\textbf{{\tool}}} &
  \multirow{2}{*}{\textbf{{QAQA}}} &
  \multirow{2}{*}{\textbf{{CAT}}} &
  \multirow{2}{*}{\textbf{SIN}} &
  \multicolumn{15}{c}{\textbf{FIP}}\\    \cline{8-22} 

 &
   &
    
    &
    &
     &
   &

   &

  \multicolumn{1}{c}{\textit{\textbf{Gaussion}}} &
  \multicolumn{1}{c}{\textit{\textbf{Shot}}} &
  \multicolumn{1}{c}{\textit{\textbf{Impulse}}} &
  \multicolumn{1}{c}{\textit{\textbf{Defocus}}} &
  \multicolumn{1}{c}{\textit{\textbf{Glass}}} &
  \multicolumn{1}{c}{\textit{\textbf{Motion}}} &
  \textit{\textbf{Zoom}} &
  \multicolumn{1}{c}{\textit{\textbf{Snow}}} &
  \multicolumn{1}{c}{\textit{\textbf{Frost}}} &
  \multicolumn{1}{c}{\textit{\textbf{Fog}}} &
  \textit{\textbf{Bright}} &
    \multicolumn{1}{c}{\textit{\textbf{Contrast}}} &
  \multicolumn{1}{c}{\textit{\textbf{Elastic}}} &
  \multicolumn{1}{c}{\textit{\textbf{Pixel}}} &
  \textit{\textbf{JPEG}} 
  \\ \midrule
\multirow{2}{*}{{RefCOCO}} &
  \textit{ACC} &
  \multicolumn{1}{c}{89.3\%} & 
  \multicolumn{1}{c}{\cellcolor{pink}\textbf{68.0\%}} & 
  \multicolumn{1}{c}{80.8\%} & 
  81.1\% & 
   
  \multicolumn{1}{c}{79.1\%} & 
  \multicolumn{1}{c}{84.8\%} &
  \multicolumn{1}{c}{84.7\%} & 85.9\% & 
  
  \multicolumn{1}{c}{85.8\%} &
  \multicolumn{1}{c}{85.1\%} &
  \multicolumn{1}{c}{85.4\%} &
  72.4\% &
   
  \multicolumn{1}{c}{81.1\%} &
  \multicolumn{1}{c}{83.9\%} &
  \multicolumn{1}{c}{87.8\%} &
  89.0\%&
   
  \multicolumn{1}{c}{83.4\%} &
  \multicolumn{1}{c}{87.5\%} &
  \multicolumn{1}{c}{88.0\%} &
  88.5\%\\ 

  &
  \textit{MMI} &
  \multicolumn{1}{c}{-} & 
  \multicolumn{1}{c}{\cellcolor{pink}\textbf{23.9\%}} & 
  \multicolumn{1}{c}{9.5\%} & 
  9.2\% & 
   
  \multicolumn{1}{c}{11.4\%} & 
  \multicolumn{1}{c}{5.0\%} &
  \multicolumn{1}{c}{5.2\%} & 3.8\% & 
  
  \multicolumn{1}{c}{3.9\%} &
  \multicolumn{1}{c}{4.7\%} &
  \multicolumn{1}{c}{4.4\%} &
  18.9\% &
   
  \multicolumn{1}{c}{9.2\%} &
  \multicolumn{1}{c}{6.0\%} &
  \multicolumn{1}{c}{1.6\%} &
  0.3\%&
   
  \multicolumn{1}{c}{6.6\%} &
  \multicolumn{1}{c}{2.0\%} &
  \multicolumn{1}{c}{1.4\%} &
  0.8\%\\

  &
  \textit{ATCR} &
  \multicolumn{1}{c}{-} & 
  \multicolumn{1}{c}{90.0\%} & 
  \multicolumn{1}{c}{86.6\%} & 
  86.0\% & 
   
  \multicolumn{1}{c}{64.0\%} & 
  \multicolumn{1}{c}{94.0\%} &
  \multicolumn{1}{c}{88.7\%} & 88.0\% & 
  
  \multicolumn{1}{c}{89.3\%} &
  \multicolumn{1}{c}{86.7\%} &
  \multicolumn{1}{c}{89.3\%} &
  67.3\% &
   
  \multicolumn{1}{c}{88.0\%} &
  \multicolumn{1}{c}{82.0\%} &
  \multicolumn{1}{c}{88.0\%} &
  92.0\%&
   
  \multicolumn{1}{c}{72.7\%} &
  \multicolumn{1}{c}{90.7\%} &
  \multicolumn{1}{c}{\cellcolor{pink}\textbf{94.7\%}} &
  94.0\%\\ 
  
   \midrule
\multirow{2}{*}{{RefCOCO+}} &
  \textit{ACC} &
  \multicolumn{1}{c}{84.4\%} & 
  \multicolumn{1}{c}{\cellcolor{pink}\textbf{64.0\%}} & 
  \multicolumn{1}{c}{78.1\%} & 
  76.2\% & 
   
  \multicolumn{1}{c}{72.0\%} & 
  \multicolumn{1}{c}{79.5\%} &
  \multicolumn{1}{c}{80.0\%} & 78.8\% & 
  
  \multicolumn{1}{c}{79.1\%} &
  \multicolumn{1}{c}{78.8\%} &
  \multicolumn{1}{c}{78.6\%} &
  74.0\% &
   
  \multicolumn{1}{c}{73.6\%} &
  \multicolumn{1}{c}{76.7\%} &
  \multicolumn{1}{c}{80.4\%} &
  83.3\%&
   
  \multicolumn{1}{c}{76.5\%} &
  \multicolumn{1}{c}{80.9\%} &
  \multicolumn{1}{c}{81.3\%} &
  81.9\%\\ 

  &
  \textit{MMI} &
  \multicolumn{1}{c}{-} & 
  \multicolumn{1}{c}{\cellcolor{pink}\textbf{24.2\%}} & 
  \multicolumn{1}{c}{7.4\%} & 
  9.7\% & 

  \multicolumn{1}{c}{14.7\%} & 
  \multicolumn{1}{c}{5.8\%} &
  \multicolumn{1}{c}{5.2\%} & 6.6\% & 
  
  \multicolumn{1}{c}{6.3\%} &
  \multicolumn{1}{c}{6.6\%} &
  \multicolumn{1}{c}{6.8\%} &
  12.3\% &
   
  \multicolumn{1}{c}{12.8\%} &
  \multicolumn{1}{c}{9.1\%} &
  \multicolumn{1}{c}{4.7\%} &
  1.3\%&
   
  \multicolumn{1}{c}{9.3\%} &
  \multicolumn{1}{c}{4.1\%} &
  \multicolumn{1}{c}{3.7\%} &
  2.9\%\\ 
  
  &
  \textit{ATCR} &
  \multicolumn{1}{c}{-} & 
  \multicolumn{1}{c}{93.3\%} & 
  \multicolumn{1}{c}{90.0\%} & 
  88.0\% & 
   
  \multicolumn{1}{c}{64.7\%} & 
  \multicolumn{1}{c}{94.0\%} &
  \multicolumn{1}{c}{91.3\%} & 94.7\% & 
  
  \multicolumn{1}{c}{92.0\%} &
  \multicolumn{1}{c}{92.7\%} &
  \multicolumn{1}{c}{90.7\%} &
  58.0\% &
   
  \multicolumn{1}{c}{90.7\%} &
  \multicolumn{1}{c}{90.7\%} &
  \multicolumn{1}{c}{92.7\%} &
  89.3\%&
   
  \multicolumn{1}{c}{80.0\%} &
  \multicolumn{1}{c}{94.0\%} &
  \multicolumn{1}{c}{94.0\%} &
  \cellcolor{pink}\textbf{96.0}\%\\ 

   \midrule
\multirow{2}{*}{{RefCOCOg}} &
  \textit{ACC} &
  \multicolumn{1}{c}{86.5\%} & 
  \multicolumn{1}{c}{\cellcolor{pink}\textbf{72.7\%}} & 
  \multicolumn{1}{c}{80.1\%} & 
  73.9\% & 
   
  \multicolumn{1}{c}{74.9\%} & 
  \multicolumn{1}{c}{79.6\%} &
  \multicolumn{1}{c}{80.5\%} & 80.1\% & 
  
  \multicolumn{1}{c}{81.0\%} &
  \multicolumn{1}{c}{80.5\%} &
  \multicolumn{1}{c}{81.9\%} &
  79.8\% &
   
  \multicolumn{1}{c}{74.8\%} &
  \multicolumn{1}{c}{77.3\%} &
  \multicolumn{1}{c}{81.1\%} &
  83.5\%&
   
  \multicolumn{1}{c}{76.9\%} &
  \multicolumn{1}{c}{80.3\%} &
  \multicolumn{1}{c}{82.1\%} &
  82.1\%\\ 

  &
  \textit{MMI} &
  \multicolumn{1}{c}{-} & 
  \multicolumn{1}{c}{\cellcolor{pink}\textbf{16.0\%}} & 
  \multicolumn{1}{c}{7.4\%} & 
  14.6\% & 

  \multicolumn{1}{c}{13.4\%} & 
  \multicolumn{1}{c}{8.0\%} &
  \multicolumn{1}{c}{6.9\%} & 7.4\% & 
  
  \multicolumn{1}{c}{6.4\%} &
  \multicolumn{1}{c}{6.9\%} &
  \multicolumn{1}{c}{5.3\%} &
  7.7\% &
   
  \multicolumn{1}{c}{13.5\%} &
  \multicolumn{1}{c}{10.6\%} &
  \multicolumn{1}{c}{6.2\%} &
  3.5\%&
   
  \multicolumn{1}{c}{11.1\%} &
  \multicolumn{1}{c}{7.2\%} &
  \multicolumn{1}{c}{5.1\%} &
  5.1\%\\ 
  
  &
  \textit{ATCR} &
  \multicolumn{1}{c}{-} & 
  \multicolumn{1}{c}{96.0\%} & 
  \multicolumn{1}{c}{95.0\%} & 
  67.0\% & 
   
  \multicolumn{1}{c}{74.2\%} & 
  \multicolumn{1}{c}{96.0\%} &
  \multicolumn{1}{c}{97.8\%} & 95.1\% & 
  
  \multicolumn{1}{c}{81.0\%} &
  \multicolumn{1}{c}{79.2\%} &
  \multicolumn{1}{c}{81.0\%} &
  58.7\% &
   
  \multicolumn{1}{c}{80.2\%} &
  \multicolumn{1}{c}{95.4\%} &
  \multicolumn{1}{c}{\cellcolor{pink}\textbf{98.2\%}} &
  98.0\% &
   
  \multicolumn{1}{c}{79.6\%} &
  \multicolumn{1}{c}{98.0\%} &
  \multicolumn{1}{c}{98.0\%} &
  96.1\%
  
   \\\bottomrule
\end{tabular}
        \end{threeparttable}
}
\end{table*}
\subsection{Answering RQ1}




Table \ref{tab:link_performance} shows the accuracy of OFA-VG model on the adversarial tests, MMI compared with performance on the original test sets, and ATCR of the adversarial tests.
In general, for the issue detection ability, the OFA-VG model shows the lowest accuracy of 64.0\%--72.7\%.
Moreover, compared with the performance on the original test sets, the adversarial tests generated by {\tool} could achieve 21.4\% in MMI, and outperform the baselines by 8.2\%--15.1\% on average.
The results indicate that {\tool} could detect issues for VG models more effectively.

Compared with the text perturbation baselines, the MMI of {\tool} is 13.3\% and 10.2\% higher than that of QAQA and CAT, respectively.
Specifically, QAQA generates tests by adding some uninformative sentences to the original expressions.
The results show that the MMI of QAQA is only 8.1\%, which indicates that compared with the perturbation techniques to increase input information like QAQA, {\tool}, which is based on property reduction, can show better issue detection performance in VG task.
For CAT, considering that it generates tests by word substitution, the MMI is also relatively low (11.2\% on average), limited by subtle perturbations.

For the quality of the adversarial tests, {\tool} still shows better performances than QAQA and CAT in ATCR.
Specifically, {\tool} is 2.6\% higher than QAQA, since QAQA may unexpectedly add the sentences, carrying the properties of other objects (rather than the target object) in the images, to the original expressions.
It leads to a mismatch between the perturbed expressions and the target objects, thus affecting the correctness of adversarial tests.
For CAT, the wrongly adversarial tests mainly result from the errors introduced by another deep model, the BERT model, used in word substitution.
In addition, since CAT replaces words following the premise of contextual semantics, there are cases where the replaced word has different semantics from the original word but fits the contextual context.
In these cases, substituted words change the descriptive properties of the expressions and produce the wrong adversarial tests.
{\tool} also introduces a deep learning model, i.e. VQA model, when implementing P1.
However, {\tool} designs three different queries from a core idea and links the three query answers with ``AND'' logic.
This ensures that a correct output is only obtained when all three queries are answered correctly, thereby offering greater confidence than using the deep learning model alone.

Compared to the image perturbation baselines, the MMI of {\tool} is 8.2\%--15.1\% higher than that of SIN and FIP.
For SIN, it perturbs images by image synthesis that is implemented by replacing the background of the original image with the stylistic features of another image. 
Although SIN shows promising performance in MMI, the ATCR of tests generated by SIN is low due to the unpredictable image synthesis.
It leads to changes in the object's properties before and after the perturbation and significantly affects the usability of the practical testing process.
As for FIP involving 15 perturbations, the results show that it is not possible to give attention to both the correct rate of test generation and issue detection ability.
Specifically, Zoom, Snow, Frost, and Contrast could achieve promising MMI (10.6\%  on average).
However, these perturbations change the background color of the image to a great extent or make the objects in the image blurred to unrecognizable, which leads to sacrificing the ATCR (78.6\% on average) of the image perturbation.
The other techniques in FIP conduct perturbations to the images within a reasonable range, resulting in higher ATCR and but less accuracy decrease.
The results also imply that compared to the expressions, the VG model shows robustness against the image perturbations.

\subsection{Answering RQ2}

Table \ref{tab:rq2} shows the MMI, i.e. the issue detection ability, after removing each perturbations.
Compared with {\tool}, ${\tool} \setminus P1$ decreases 6.3\%--8.3\% in MMI on the three datasets.
The results indicate that the property reduction could obviously improve the issue detection ability of {\tool}.
Compared with {\tool}, ${\tool} \setminus P2$ decreases 9.3\%--11.4\% in MMI on the three datasets.
The results indicate that semantically-equivalent text perturbations are still effective means for test generation for the VG models.
In particular, {\tool} (i.e., combining P1 and P2) achieves an MMI of 16.0\%--24.2\% on the three datasets.
This performance significantly surpasses that of a single perturbation and even exceeds the combined MMI of two perturbations on the RefCOCOg dataset.
The two perturbations are designed from different aspects of natural language.
The results show a more effective issue detection ability when combined with the two perturbations in {\tool}.

\begin{table}[h]\huge
\caption{The MMI of {\tool} after removing different perturbations}
\label{tab:rq2}
\resizebox{\columnwidth}{!}{
\begin{tabular}{c|ccc}
\toprule
\textbf{Experiment Group} & \textbf{RefCOCO} & \textbf{RefCOCO+} & \textbf{RefCOCOg} \\ \midrule
{\tool} &   23.9\%     &   24.2\%   &  16.0\% \\ 

${\tool} \setminus P1$  & 17.6\% (-6.3\%)  & 17.4\% (-6.8\%)  &   7.7\% (-8.3\%)   \\ 

${\tool} \setminus P2$  &   14.6\% (-9.3\%)    & 14.2\% (-10.0\%) &   4.6\% (-11.4\%)    \\


\bottomrule
\end{tabular}
}
\end{table}

\subsection{Answering RQ3}


In this RQ, we investigate whether the adversarial tests generated by {\tool} can aid in improving the performance of the VG model. 
For each dataset, we divide the tests generated by {\tool} into a training set and an evaluation set following the experimental settings described in Section \ref{sec:experiment_design}. 
Given the OFA-VG model, we use the training set to fine-tune the model and the new evaluation set to evaluate its performance. 
Figure \ref{tab:rq3} displays the accuracy of both the original OFA-VG model and the fine-tuned OFA-VG model on the evaluation set. 
The results indicate that the fine-tuned OFA-VG model with the adversarial tests can indeed improve the accuracy of the VG model. 
Notably, the model's accuracy on the RefCOCO+ dataset increases significantly by up to 35.8\%. 
The evaluation results imply that the tests generated by {\tool} not only detect issues in the VG model but also enhance the original VG model.
Moreover, by identifying the model's issues using {\tool}, developers can focus on refining specific aspects of the model, leading to a more accurate and reliable VG model. 
In general, the adversarial tests generated by {\tool} serve as valuable resources for model development and enhancement, contributing to the advancement of the VG task.

\begin{table}[h] \huge
\caption{Improved accuracy after fine-tuning with tests generate by {\tool}}
\label{tab:rq3}
\resizebox{\columnwidth}{!}{
\begin{tabular}{c|ccc}
\toprule
\textbf{Test Model} & \textbf{RefCOCO} & \textbf{RefCOCO+} & \textbf{RefCOCOg} \\ \midrule

Original model &   68.1\%     &   64.0\%     &   72.9\%     \\

Fine-tuned model  & 80.5\% ($\uparrow 18.2\%$)  & 86.9\% ($\uparrow 35.8\%$) &   91.8\% ($\uparrow 25.9\%$)   \\

\bottomrule
\end{tabular}
}
\end{table}

\subsection{Answering RQ4}
\begin{table}[t]\scriptsize
\caption{ Object and properties extraction performance}
\label{tab:entity_extraction}
\centering
\resizebox{\columnwidth}{!}{
\begin{tabular}{cc|cccc}
\toprule
\textbf{Category}  & \textbf{Model}  & \textbf{Precision} & \textbf{Recall} & \textbf{F1} \\ \midrule
\multirow{2}{*}{Object} & Tscope & 94.9\%  & 93.8\%  & 94.3\%     \\
  & {\tool} & \textbf{96.0\%}  & \textbf{96.0\%}  & \textbf{96.0\%}     \\ \midrule
\multirow{2}{*}{Property}&  Tscope& 91.3\%  & 74.8\%  & 82.2\%    \\

&  {\tool}& \textbf{95.7\%}  & \textbf{94.0\%}  & \textbf{94.8\%}    \\

\bottomrule
\end{tabular}
}
\end{table}
Table \ref{tab:entity_extraction} shows the performance of object and properties extraction. 
In general, {\tool} could achieve promising extraction performance with 95.9\% precision, 95.0\% recall, and 95.4\% F1.
Compared to Tscope, which is a supervised information extraction baseline, the overall performance (F1) of {\tool} for the object and properties surpasses by 1.7\% and 12.6\% respectively.
Owing to the diverse representations of properties (e.g., action, mood, color, wear, location, behavior, etc.), it is challenging for Tscope to fully capture these properties with the limited labeled data.
However, LLM such as ChatGPT has already learned general knowledge from vast amounts of data.
Hence, it is more generalizable and can accurately capture property features by relying on in-context learning with only a small amount of labeled data.
Therefore, {\tool} requires far less labeled data than Tscope, which greatly saves labor and time costs.

\section{Discussion}
\label{sec:discussion}

\subsection{The Detected Issues in VG Model}

\begin{figure*}[h]
  \center{\includegraphics[width=\linewidth]{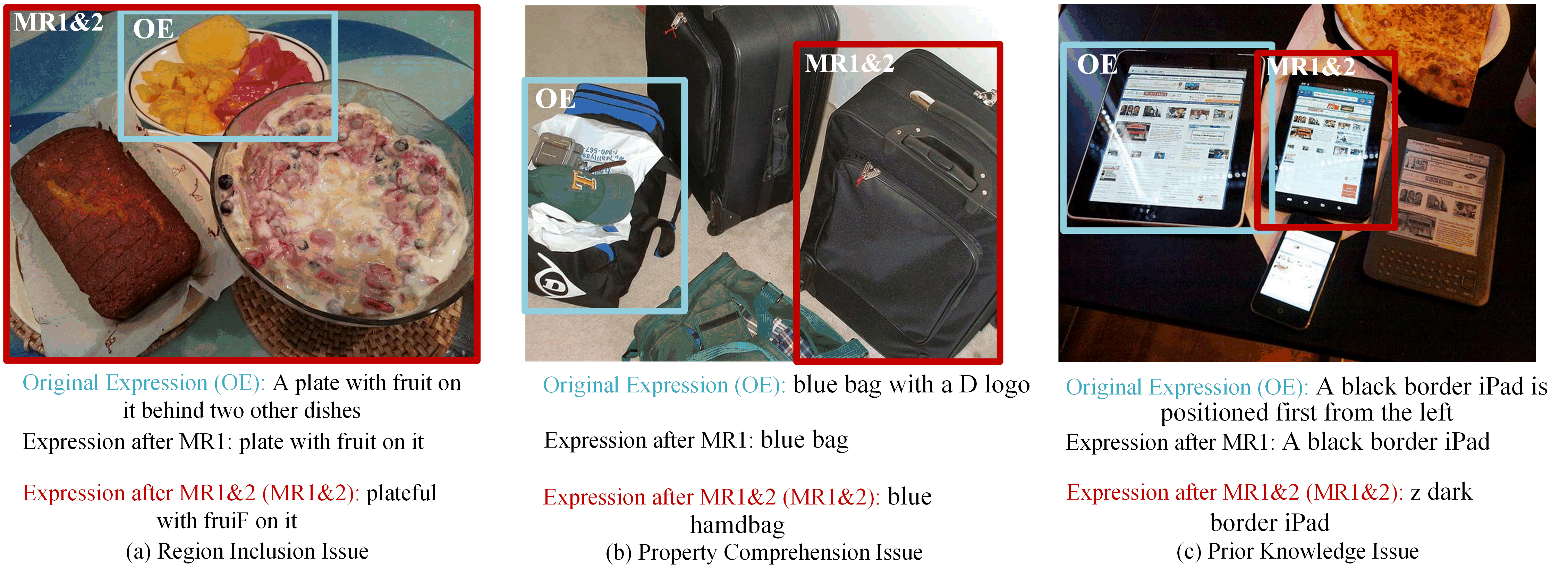}}
    \caption{
     Three categories of detected issues by {\tool} in VG testing
    }
    \label{fig:badcase}    
\end{figure*}

This section demonstrates three typical categories of the detected issues by {\tool}, to facilitate the follow-up studies in improving the VG models. 
In Figure \ref{fig:badcase}, for each issue, we show the image, the original expression (OE), the expression perturbed by the single P1, and the expression perturbed by both P1\&2.

The first type of detected issue is related to the region inclusion problem.
Specifically, as shown in subfigure (a) of Figure \ref{fig:badcase}, the expression after P1\&2 makes the VG model locate the region of the whole image, which encompasses the original target region. 
This indicates that the scope of the model's localized region needs further refinement.
The second detected issue type is relevant to the ability to capture the detailed properties in the expression.
Specifically, as shown in subfigure (b) of Figure \ref{fig:badcase}, the VG model does not capture the complete semantics of the expression (i.e., ``blue'' in subfigure (b) is overlooked by the model) after property reduction and results in the incorrect predicted region. 
This indicates that the VG model needs to enhance the ability to capture the detailed properties in expressions.
The third type of detected issue is the problem of lacking general prior knowledge.
Specifically, as shown in subfigure (c) of Figure \ref{fig:badcase}, upon incorporating external knowledge, the physical characteristics of an iPad typically include a round button at the bottom, and devices that have four operating keys at the bottom are usually Android devices.
 As the VG model lacks prior information derived from external knowledge, it is hard to locate the target region with the limited information available.
 This suggests that incorporating external knowledge in the design of VG models is a potential way to enhance their performance.

\subsection{The Credibility of the VQA Model}
\label{sec:discussion_confidence}

When performing property reduction, {\tool} employs the VQA model to validate each candidate property reduction expression.
Therefore, the errors in the VQA model may accumulate in the tests generated by {\tool} and affect their quality.
To this end, we discuss the credibility of the VQA model from three aspects.
First, as reported by the previous studies~\cite{wang2022ofa}, the VQA model is a state-of-the-art model and has achieved an accuracy of 94.6\% for the yes/no queries (the same as the query type in {\tool}).
Its admirable performance is a guarantee of credibility that employs it in {\tool}.
Second, to further ensure the credibility of the VQA outputs, {\tool} leverages three queries that obtain the answers from three individual views.
After that, {\tool} comprehensively considers the answers to three queries and validates each candidate property reduction expression.
This multilateral mechanism can effectively increase the credibility of VQA output results.
Third, according to the correct rate of the adversarial tests, the results reveal that the average ATCR is 92.1\%, which implies that errors in VQA do not have a significant impact on the adversarial tests.




\subsection{Analysis of Image Perturbations in VG Task}
\label{sec:analysis_img}

\begin{figure}[h]
\centering
\includegraphics[width=\linewidth]{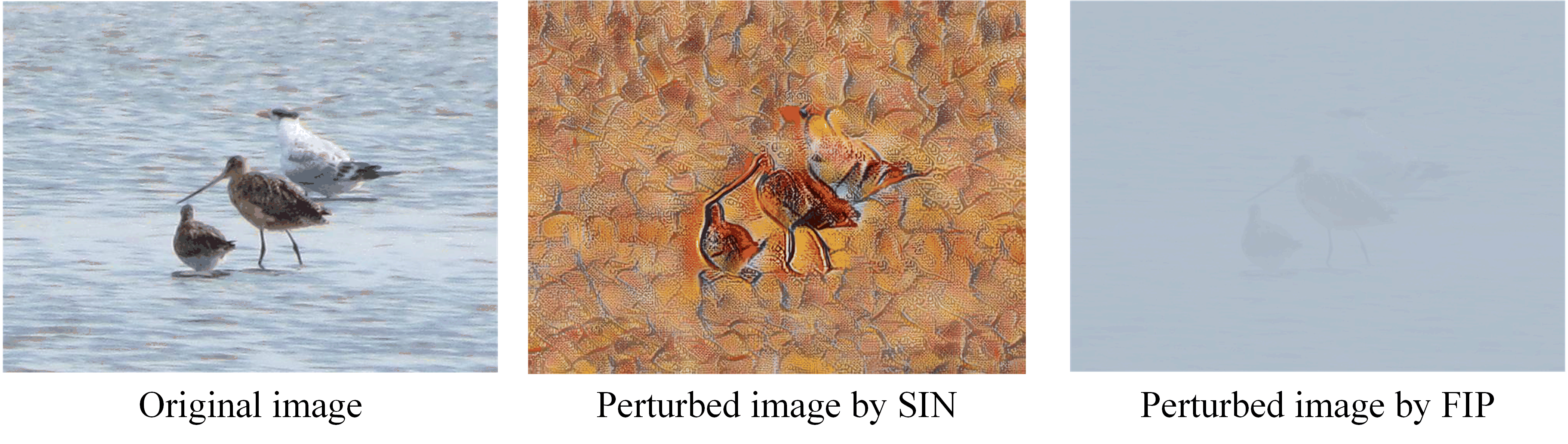}
\caption{ An original image and the images perturbed by SIN and FIP respectively}
\label{fig:img_badcase}
\end{figure}

In this section, we use an example to illustrate why image perturbation cannot be effectively used for test generation.
Figure \ref{fig:img_badcase} gives an original image for the VG task that locates ``A white bird standing behind two brown birds'' as well as two perturbed images using SIN and FIP respectively.

Both SIN and FIP add a certain amount of noise to all the pixels in the images. Since the location and degree of noise cannot be accurately estimated, SIN and FIP could not determine whether the perturbed images could make the test oracles fail.
According to the image, it can be seen that the expression describes two discriminative properties, i.e., a color property ``white'' and a position property ``stands behind two brown birds''.
For the image perturbed by SIN, SIN leads to obvious changes to the properties of the object in the image, and even it is difficult to distinguish the types of the target object.
As a consequence, the target object in the perturbed image could not be located according to the original expression describing the color feature.
Not limited to two perturbations in the above examples, previous image perturbations are typically implemented following similar technical principles.
The randomness and inexplicability of these techniques lead to the inability to obtain expected perturbed results, and therefore they cannot be effectively applied to test generation.

\subsection{Threats to Validity}
\textbf{{External Validity}}.
The external threats are related to the generalization of the proposed approach.
First, we experimented with the data taken from three open datasets, and the results may vary on other datasets.
Second, {\tool} requires perturbing the text after integrating information from both the image and the expression, making it inapplicable to single-mode deep learning models.
Third, {\tool} is required to extract objects and properties from the expressions, which may not be suitable for different multimodal models.

\textbf{{Internal Validity}}. 
The internal threats relate to experimental errors and biases.
First, we did not evaluate the ATCR for the whole set of adversarial expressions since automatically calculating the ATCR can be error-prone.
However, we randomly sample three sets of data and take the average of their ATCRs to represent the performance of the full set, which is widely-used practice in many studies \cite{SwamyS07,Ma0WX13} and could alleviate the threat.
Second, employing the VQA model to determine whether the candidate property reduction expression only describes the original target region in the image may introduce bias resulting from the VQA model's inaccurate judgment.
We designed the queries from three perspectives and only when the three queries receive the expected answer, we consider the candidate property reduction expression as the satisfactory one, which may alleviate the threat.

\section{Related Work}
\label{sec:related work}


In this section, we will discuss the related works in two aspects, i.e., adversarial testing for NLP models and adversarial testing for CV models under the framework of black box.
\subsection{Adversarial Testing for NLP Models}
As NLP technology is rapidly developing, numerous methods have been proposed to test NLP models. 
On the one hand, researchers have proposed some unsupervised approaches. For example, Liang et al. \cite{0002LSBLS18} design three perturbation strategies, namely insertion, modification, and removal, to generate adversarial samples and keep semantic equivalence. Liu et al. \cite{LiuF021} employ a series of transformation operators to make realistic changes to seed data while preserving their oracle information properly. Moreover, many researchers \cite{SunZHPZ20,0004Z0HP022,RusertSS22} propose various strategies for substituting synonymous words in the original test input. Regarding perturbation strategies, existing approaches typically perturb the text at the character, word, or sentence level while preserving semantic equivalence. These methods, however, only offer a shallow level of perturbation, which consequently impacts the issue detection ability.

On the other hand, many studies focus on adversarial attacks, which can be classified into two categories, i.e., non-targeted attacks and targeted attacks. Non-targeted attacks are carried out by modifying some important words that can affect the classification results. For example, Papernot et al. \cite{PapernotMSH16} first study adversarial examples in the text domain by exploiting computational graph expansions to evaluate forward derivatives, which are associated with word vectors. Then they use the results calculated by FGSM to find out the anti-disturbance. Sato et al. \cite{SatoSS018} propose a method for operating in the input word vector space, called iAdv-Text. Its core process can be regarded as an optimization problem. 
As for targeted attacks, Ebrahimi et al. \cite{EbrahimiRLD18} propose a method called HotFlip, which is similar to the one-pixel attack. HotFlip is a white-box attack method that relies on an atomic flip operation to swap characters through gradient computation. 
Gil et al. \cite{GilCGB19} propose a derived method, DISTFLIP, which distills knowledge from HotFlip and uses it to train black-box models. By training the model, the study generates adversarial examples for black-box attacks. Though those approaches are useful and effective in NLP testing, they are not appropriate for VG testing. This kind of approach depends on annotated text data, which is more costly in VG because we need to consider both visual information and text information in VG annotation.



\subsection{Adversarial Testing for CV Models}
In the literature, several approaches have been proposed to test and improve computer vision models. Those approaches can be classified into unsupervised approaches and supervised approaches. Unsupervised approaches focus on perturbing images in different aspects. For example, Dan et al. \cite{HendrycksD19} assess performance on common corruptions and perturbations by creating corrupted images using various techniques, including noise, blur, weather-related effects, and digital manipulations. To test Vision Transformer (ViT), Bhojanapalli et al. \cite{BhojanapalliCGL21} apply both input perturbations and model perturbations. Paul et al. \cite{PaulC22} study the robustness of the Vision Transformer against common corruptions and perturbations, distribution shifts, and natural adversarial examples using six different diverse ImageNet datasets concerning robust classification. 
In addition, some supervised approaches that mainly focus on adversarial attacks are also proposed to test CV models. As for adversarial attack methods, the first gradient-based attack method is the fast gradient sign method (FGSM), proposed by Goodfellow et al. \cite{GoodfellowSS14}. FGSM adds perturbations along the opposite direction of the gradient to increase the loss function rapidly, which eventually leads to model classification errors. At present, researchers generally believe that the projected gradient descent method (PGD) \cite{MadryMSTV18} is the best gradient-based attack. PGD is essentially an improvement of the I-FGSM method. By adding a layer of randomization processing, the number of iterations is increased, and the attack is greatly improved. Inspired by the concept of saliency maps, Papernot et al. \cite{PapernotMJFCS16} propose a Jacobian-based Matrix saliency map attack method (JSMA). Specifically, they first use the gradient information to calculate the pixel position that has the greatest impact on the classification result, and then add perturbation to the pixel to obtain an adversarial example.

To evaluate the robustness of CV models, most approaches typically involve perturbing image information or generating adversarial examples.
However, these approaches are not suited for testing VG models. Techniques for CV testing focus solely on perturbing visual information, neglecting the vital role of text information in VG tasks. Moreover, perturbing the image in the VG task could alter its content, leading to incompatibility between the original expression and the newly generated image.
Adversarial attacks on CV models are also not suitable for the VG task, as these supervised methods are specifically tailored for the object detection task that deals with categorizing images.

\section{Conclusion}
\label{sec:conclusion}
As a fundamental task in multimodal learning, the VG task has been widely-used in our daily scenarios, but the corresponding quality issues are still serious in applications.
In this paper, we present {\tool}, a text perturbation approach via image-aware property reduction for adversarial testing of the VG model.
Besides the prevalent semantically-equivalent perturbations, {\tool} designs a novel property reduction perturbation for expressions.

To implement the novel expression perturbation, {\tool} extracts the object and properties from the original expression and recombines them into candidate property reduction expressions.
Then, it designs three queries for each candidate property reduction expression. 
The queries and images are fed into the VQA model to select the accurate expressions for only describing the original object.
The evaluation shows {\tool} could outperform the state-of-the-art baselines for the VG task.
In addition, by fine-tuning the original VG model with the adversarial tests generated by the {\tool}, the performance of the VG model can be significantly improved.
In the future, we will further resolve the issues of the VG model exposed by the {\tool} and investigate the effectiveness of novel perturbations in other multimodal tasks.


\bibliographystyle{IEEEtran}
\bibliography{ref}

\vfill

\end{document}